\documentclass[a4paper]{article}

\usepackage{times}
\usepackage{url}
\usepackage{amsmath,amssymb,amsthm}
\usepackage{graphicx}
\usepackage{comment}
\usepackage{fullpage}
\usepackage{subfig}
\usepackage[round]{natbib}
\usepackage[T1]{fontenc}
\usepackage[utf8]{inputenc}

\newcommand{\x}{\mathbf{x}}
\newcommand{\xs}{\mathbf{x}_\star}
\newcommand{\y}{\mathbf{y}}
\newcommand{\f}{\mathbf{f}}
\newcommand{\fs}{\mathbf{f}_\star}
\renewcommand{\l}{\ell}
\newcommand{\bl}{\boldsymbol\ell}
\newcommand{\s}{\sigma}
\newcommand{\balpha}{\boldsymbol\alpha}
\newcommand{\0}{\mathbf{0}}
\newcommand{\bs}{\boldsymbol\sigma}
\newcommand{\bmu}{\boldsymbol\mu}
\newcommand{\e}{\varepsilon}

\newcommand{\bt}{\boldsymbol\theta}
\renewcommand{\o}{\omega}
\newcommand{\bo}{\boldsymbol\omega}
\newcommand{\N}{\mathcal{N}}
\newcommand{\R}{\mathbb{R}}

\renewcommand{\L}{\mathcal{L}}

\newcommand{\ts}{{\tilde{\sigma}}}
\newcommand{\tl}{{\tilde{\ell}}}
\renewcommand{\to}{{\tilde{\omega}}}
\newcommand{\tbl}{{\tilde{\boldsymbol\ell}}}
\newcommand{\tbs}{{\tilde{\boldsymbol\sigma}}}
\newcommand{\tbo}{{\tilde{\boldsymbol\omega}}}
\newcommand{\bbl}{{\mathring{\boldsymbol\ell}}}
\newcommand{\bbs}{{\mathring{\boldsymbol\sigma}}}
\newcommand{\bbo}{{\mathring{\boldsymbol\omega}}}
\newcommand{\map}{{\textsc{map}}}

\DeclareMathOperator{\diag}{diag}
\DeclareMathOperator{\tr}{tr}

\DeclareMathOperator{\cov}{\textbf{cov}}

\DeclareMathOperator*{\argmax}{arg\,max}

\title{Non-Stationary Gaussian Process Regression with Hamiltonian Monte Carlo}

\author{Markus Heinonen$^{1,2}$\footnote{Corresponding author, \texttt{markus.o.heinonen@aalto.fi}}\,\,, Henrik Mannerstr{\"o}m$^{2}$, Juho Rousu$^{1,2}$, Samuel Kaski$^{1,2}$ and Harri L{\"a}hdesm{\"a}ki$^{2}$ \\
\\
$^1$Helsinki Institute for Information Technology HIIT, Finland\\
$^2$Department of Computer Science, Aalto University, Finland\\}

\begin{document}

\maketitle

\begin{abstract}
We present a novel approach for fully non-stationary Gaussian process regression (GPR), where all three key parameters -- noise variance, signal variance and lengthscale -- can be simultaneously input-dependent. We develop gradient-based inference methods to learn the unknown function and the non-stationary model parameters, without requiring any model approximations. We propose to infer full parameter posterior with Hamiltonian Monte Carlo (HMC), which conveniently extends the analytical gradient-based GPR learning by guiding the sampling with model gradients. We also learn the MAP solution from the posterior by gradient ascent. In experiments on several synthetic datasets and in modelling of temporal gene expression, the nonstationary GPR is shown to be necessary for modeling realistic input-dependent dynamics, while it performs comparably to conventional stationary or previous non-stationary GPR models otherwise.
\end{abstract}

\section{Introduction}

Gaussian process regression has emerged as a powerful, yet practical class of non-parametric Bayesian models that quantify the uncertainties of the underlying process using Gaussian distributions \citep{rasmussen06}. Gaussian processes are commonly applied to time-series interpolation, regression and classification, where the GP can provide predictive distributions \citep{rasmussen06}. 

The standard GP model assumes that the model parameters stay constant over the input space. This includes the observational noise variance $\o^2$, as well as the signal variance $\s^2$ and the lengthscale $\l$ of the covariance function. The signal variance determines the signal amplitude, while the characteristic lengthscale defines the local `support' neighborhood of the function. In many real world problems either the noise variance or the signal smoothness, or both, vary over the input space, implying a \emph{heteroscedatic} noise model or a \emph{nonstationary} function dynamics, respectively \citep{le05} (see also \citet{wang12}). In both cases, the analytical posterior of the GP becomes intractable \citep{tolvanen14}. For instance, in biological studies, rapid signal changes are often observed quickly after perturbations, with the signal becoming smoother in time \citep{heinonen15}. 

Several authors have proposed extending GPs by learning a latent noise variance as another GP, and by inferring the unknown function and the noise model in a maximum likelihood (ML) \citep{kersting07} or maximum a posteriori (MAP) fashion \citep{quadrianto09}. Fully Bayesian inference methods include MCMC sampling \citep{goldberg98} and variational and expectation propagation approximations of the posterior \citep{lazaro11,tolvanen14}. Non-stationarities can also be included in the signal variance or lengthscale with Gaussian process priors. Nonstationary lengthscale was introduced by \citet{gibbs97} and further extended by \citet{paciorek04} with MCMC inference. Recently, \citet{tolvanen14} introduced a non-stationary signal variance using expectation propagation and approximate variational inference.

In this paper we introduce the first fully non-stationary and heteroscedastic GP regression framework, in which all three main components (noise variance, signal variance and the lengthscale) can be simultaneously input-dependent, with GP priors\footnote{Matlab implementation available from \url {github.com/markusheinonen/adaptivegp}}. We propose an inference method for the exact joint posterior of the underlying signal and all three latent functions, avoiding the need for introducing variational or expectation propagation approximations \citep{lazaro11,tolvanen14}. We use HMC-NUTS, which can effectively sample the posterior guided by the model gradients, which we derive analytically. Furthermore, an exact MAP solution arises as a simple gradient ascent of the posterior. We enhance both approaches by posterior whitening using Cholesky decompositions of the latent function priors. Our experiments demonstrate the necessity of non-stationary GPR to model realistic input-dependent dynamics, while the proposed method performs comparably to conventional stationary or previous non-stationary GPR models otherwise.

In Section \ref{sec:gp} we introduce the fully nonstationary GP model. In its subsections we first introduce MAP and HMC inference, discuss model whitening and finally define the predictive distributions. Section \ref{sec:exp} presents experimental results on several synthetic and one real biological datasets, and we conclude in Section \ref{sec:conclusions}.

%
\section{Heteroscedatic nonstationary GP model}
\label{sec:gp}

Let $\y = (y_i)_{i=1}^n \in \R^n$ be an observation vector over $n$ inputs $\x = (x_i)_{i=1}^n \in \R^n$. We assume an additive regression model 
\begin{align*}
y(x) = f(x) + \e(x), \quad \e(x) \sim \N(0, \o(x)^2),
\end{align*}
where both the underlying signal $f(x)$ and the zero-mean observation noise variance $\o(x)^2$ are unknown functions to be learned. We proceed by first placing a zero mean GP prior on the unknown function $f(x)$,
\begin{align}
f(x) \sim GP(0, K_f(x,x')),
\end{align}
which assumes that the output covariances  $\cov( f(x), f(x') ) = K_f(x, x')$ depend on the input covariance through a kernel function. We use a nonstationary generalisation of the squared exponential kernel \citep{gibbs97}
\begin{align} \label{eq:nskernel}
K_f(x,x') = \sigma(x) \sigma(x')  \sqrt{\frac{2 \l(x) \l(x')}{ \l(x)^2 + \l(x')^2 }} \exp \left( - \frac{(x - x')^2}{\l(x)^2 + \l(x')^2}  \right),
\end{align}
where $x, x' \in \R$, and $\sigma(x)$ and $\l(x)$ are input-dependent signal variance and lengthscale functions, respectively. The kernel reduces into a standard squared exponential kernel if both are constant. We show the kernel \eqref{eq:nskernel} is positive definite in the Supplementary Material.

We model the lengthscale, signal variance and noise variance with \emph{latent functions}. We are interested in smoothly varying latent functions and thus we place separate GP priors on them as well:
\begin{align*}
\log(\l(t)) \equiv \tl(t) &\sim GP(\mu_\l, K_\l(x,x')) \\
\log(\s(t)) \equiv\ts(t) &\sim GP(\mu_\s, K_\s(x,x'))\\
\log(\o(t)) \equiv \to(t) &\sim GP(\mu_\o, K_\o(x,x')), 
\end{align*}
where we set the priors on the logarithms to ensure their positivity. We select separate standard squared exponential covariances for each, $$K_c(x,x') = \alpha_c^2 \exp\left(- \frac{(x-x')^2}{2 \beta_c^2} \right),$$ where $c \in \{\l, \s, \o\}$. The model has nine hyper-parameters $\bt = (\mu_\ell, \mu_\sigma, \mu_\o, \alpha_\ell, \alpha_\sigma, \alpha_\o, \beta_\ell, \beta_\sigma, \beta_\o)$ that define the prior for the three latent functions $\tl$, $\ts$ and $\to$. The means $\mu$ determine latent function means, while the $\alpha$'s are scaling terms. The $\beta$'s are the characteristic lengthscales of the priors. 
In practice, the $\mu$'s and $\alpha$'s have a small effect on the models, whereas $\beta$'s determine the smoothness of the latent functions, and can be set based on prior knowledge.

Given a dataset $(\x,\y)$, the model can be equivalently written as $\f | \bl,\bs \sim \N(\0, K_f)$, where $\f = (f(x_i))_{i=1}^n$ is a latent function vector at observed points $\x$ and $K_f \in \R^{n \times n}$ has elements $[K_f]_{ij} = K_f(x_i,x_j)$ computed using eq.~\eqref{eq:nskernel} with signal variances $\bs = (\s(x_i))_{i=1}^n$ and lengthscales $\bl = (\l(x_i))_{i=1}^n$. Finally, the data likelihood is $\y | \bl,\bs \sim \N(\0, K_f + \Omega)$, where $\Omega = \diag \bo^2 \in \R^{n \times n}$ is a diagonal noise matrix and $\bo = (\o(x_i))_{i=1}^n$ are the noise standard deviations.

We note that by placing a GP prior on just the noise and restricting the other two to be constants, we arrive at the heteroscedastic model studied in several earlier works \citep{goldberg98,kersting07,quadrianto09}. Setting a prior on only the lengthscale retrieves the models of \citet{gibbs97,paciorek04}, and setting a prior on both the signal variance and the noise gives the model of \citet{tolvanen14}. Out method is the first to combine heteroscedatic noise and nonstationary lengthscale, while also allowing the signal variance to vary over the inputs.

To infer latent functions from the \emph{full posterior} $p(\f, \tbl, \tbs, \tbo | \y, \bt)$ we introduce two approaches in the next two Sections\footnote{In the following we omit the hyperparameters $\bt$ for notational clarity}. We  propose to learn the MAP estimate $p(\f | \tbl_\map, \tbs_\map, \tbo_\map , \y)$, or infer the full posterior using HMC sampling. Both approaches are based on the analytical gradients of the latent functions.


\subsection{Maximum a posteriori estimation}
\label{sec:map}

As the first approach, we follow the approaches by \citet{kersting07} and \citet{quadrianto09}, and resort to finding the MAP solution of the \emph{latent posterior} $p(\tbl, \tbs, \tbo | \y)$,
\begin{align*}
\tbl_\map,\tbs_\map, \tbo_\map &= \argmax_{\tbl, \tbs, \tbo} p(\tbl, \tbs, \tbo |\y),
\end{align*}
where $\f$ has been marginalised out. Using Bayes' theorem this is equivalent to maximizing the marginal likelihood 
\begin{align} \label{eq:L}
\L = p(\y | \tbl, \tbs, \tbo) p(\tbl, \tbs, \tbo) &= \N(\y |\0,  K_f + \Omega)  \N(\tbl | \mu_\ell, K_\l)  \N(\tbs | \mu_\sigma, K_\s) \N(\tbo | \mu_\o, K_\o),
\end{align}
whose logarithm we denote as the marginal log likelihood (MLL).

The partial derivatives of the log of marginal likelihood \eqref{eq:L} with respect to the latent functions are analytical:
\begin{align} \label{eq:grads}
\frac{\partial \log \L}{\partial \tbl_i }  &= 0.5 \tr \left( (\balpha \balpha^T - K_y^{-1}) \frac{\partial K_y}{\partial \tbl_i} \right) - [K_{\tl}^{-1} (\tbl - \mu_{\tl})]_i  \nonumber \\
\frac{\partial \log \L}{\partial \tbs } &= \diag \left( (\balpha \balpha^T - K_y^{-1}) K_f \right) - K_{\ts}^{-1} (\tbs - \mu_{\ts})  \\
\frac{\partial \log \L}{\partial \tbo }  &= \diag \left( (\balpha \balpha^T - K_y^{-1}) \Omega \right) - K_{\to}^{-1} (\tbo - \mu_{\to}) \nonumber
\end{align}	
where $\balpha = (K_f + \Omega)^{-1} \y$ and $\frac{\partial K_y}{\partial \tbl_i}$ is given in the Supplementary Material. 

We perform gradient ascent over the MLL, $\log \L$. The solution is only guaranteed to converge to a local optimum, and hence we perform multiple restarts from random initial conditions. The MAP solution is adequate when the posterior is close to unimodal.

Given the MAP solution, the function posterior $p(\f | \tbl_\map, \tbs_\map, \tbo_\map) \sim \N(\bmu_\map, \Sigma_\map)$ is a Gaussian with
\begin{align*}
\bmu_\map &= K_f^T (K_f + \Omega_\map)^{-1} \y \\
\Sigma_\map &= K_f - K_f^T (K_f + \Omega_\map)^{-1} K_f,
\end{align*}
where $K_f$ has been computed with eq.~\eqref{eq:nskernel} using MAP latent vectors $\log(\bl) = \tbl_\map$ and $\log(\bs) = \tbs_\map$, and $\Omega_\map$ with $\log(\bo) = \tbo_\map$.


\subsection{HMC inference}
\label{sec:hmc}

As a second approach we sample the full posterior using Hamiltonian Monte Carlo (HMC) \citep{hoffman14,neal11}. In HMC, 
we introduce an additional momentum variable for each of the model variables and interpret the extended model as a Hamiltonian system. We simulate time evolution of the Hamiltonian dynamics to produce proposals for the Metropolis algorithm. This simulation step makes use of the gradient of the log joint density
\begin{align*}
p(\f, \tbl, \tbs, \tbo; \y) &= p(\y | \f, \tbo) p(\f | \tbl, \tbs) p(\tbl | \theta) p(\tbs | \theta) p(\tbo | \theta). 
\end{align*}
However, in a GP regression the \emph{function posterior} $p(\f | \y)$ with latent functions $\tbl,\tbs,\tbo$ marginalised out, can be integrated conveniently by
\begin{align}
p(\f | \y) &= \iiint p(\f | \tbl, \tbs, \tbo, \y) p(\tbl, \tbs, \tbo | \y) d\tbl d\tbs d\tbo \\
             &\approx  \frac{1}{m} \sum_{i=1}^m p(\f | \tbl_i, \tbs_i, \tbo_i , \y), \nonumber
\end{align}
where
\begin{align}
\tbl_i, \tbs_i, \tbo_i &\sim p(\tbl, \tbs, \tbo | \y)
\end{align}
are $m$ HMC samples of the latent posterior. The function posterior $p(\f | \tbl_i, \tbs_i, \tbo_i , \y) = \N(\bmu_i, \Sigma_i)$ for each HMC sample is a Gaussian with
\begin{align*}
\bmu_i &= K_{f_i}^T (K_{f_i} + \Omega_i)^{-1} \y \\
\Sigma_i &= K_{f_i} - K_{f_i}^T (K_{f_i} + \Omega_i)^{-1} K_{f_i},
\end{align*}
where $K_{f_i}$ is a nonstationary kernel matrix computed using $\tbl_i$ and $\tbs_i$, and $\Omega_i$ is the diagonal noise covariance matrix of  $\tbo_i$. 

Hence, we only need to do HMC sampling over the three latent vectors $(\tbl, \tbs, \tbo)$ and the posterior of $\f$ follows analytically as a mixture of $m$ Gaussians. The latent posterior $p(\tbl, \tbs, \tbo | \y)$ is proportional to the marginal likelihood in eq.~\eqref{eq:L}, and thus the HMC sampling of $(\tbl, \tbs, \tbo)$ uses the same gradients $\left(\frac{\partial \log \L}{\partial \tbl}, \frac{\partial \log \L}{\partial \tbs}, \frac{\partial \log \L}{\partial \tbo}\right)$ from eq.~\eqref{eq:grads} as the MAP solution. 

\subsection{Posterior whitening}
\label{sec:white}

The posterior of the latent vectors is by definition highly correlated due to Gaussian priors, leading to inefficient Monte Carlo sampling. To ease the sampling, we perform the sampling over the whitened latent vectors \citep{kuss05}
\begin{align*}
\bbl &= L_\l^{-1} \tbl, \quad K_\l = L_\l L_\l^T \\
\bbs &= L_\s^{-1} \tbs, \quad K_\s = L_\s L_\s^T  \\
\bbo &= L_\o^{-1} \tbo, \quad K_\o = L_\o L_\o^T,
\end{align*}
with Cholesky decompositions of the corresponding GP prior covariances, which are fixed based on the hyperparameters $\bt$. The derivative for e.g. the lengthscale becomes $\frac{\partial \log \L }{\partial \bbl} = \frac{\partial \log \L }{\partial L_\l \bbl} \frac{\partial L_\l \bbl }{\partial \bbl} =  L_\l^T \nabla_{\tbl} \L$, where the last term is the standard gradient of the non-whitened model defined in eq.~\eqref{eq:grads}.

\subsection{Making predictions}
\label{sec:pred}

Both the MAP solution and the HMC sampler infer values of the latent functions only at the $n$ observed inputs $\x$. To extrapolate the values of the unknown function and the latent functions over arbitrary target points $\xs \in \R^{n_\star}$, we define the \emph{predictive distribution}, where we extrapolate the latent functions $\tbl,\tbs,\tbo$ to $\tbl_\star,\tbs_\star,\tbo_\star$, and then express the function posterior $\fs$ with them \citep{goldberg98}. With the MAP solution we have
\begin{align} \label{eq:predmap}
p(\fs | \tbl_\map, \tbs_\map, \tbo_\map, \y) &= \iint p(\fs | \tbl_\map, \tbl_\star, \tbs_\map, \tbs_\star, \tbo_\map, \y) p(\tbl_\star | \tbl_\map)  p(\tbs_\star | \tbs_\map)  d\tbl_\star d\tbs_\star \nonumber \\
   &\approx \frac{1}{s} \sum_{j=1}^s p(\fs | \tbl_\map, \tbl_{j_\star}, \tbs_\map, \tbs_{j_\star}, \tbo_\map, \y)
\end{align}
where we approximate the integral by drawing $s$ samples  $\{ \tbl_{j_\star} \}_{j=1}^s$, $\{ \tbs_{j_\star} \}_{j=1}^s$ of $n_\star$ dimensions from the conditional Gaussians $\tbl_\star | \tbl_\map$ and $\tbs_\star | \tbs_\map$ (See Supplementary Material). This results in a mixture of $s$ corresponding Gaussians $\N(\bmu_{\map,{j_\star}}, \Sigma_{\map,{j_\star}})$, where
\begin{align*}
\bmu_{\map,{j_\star}}    &= K_{\map,{j_\star}}^T( K_{\map,\map} + \Omega_\map  )^{-1} \y \\
\Sigma_{\map,{j_\star}} &= K_{{j_\star},{j_\star}} -  K_{\map,{j_\star}}^T( K_{\map,\map} + \Omega_\map  )^{-1} K_{\map,{j_\star}},
\end{align*}
and where $K_{{j_\star},{j_\star}} \in \R^{n_\star \times n_\star}$, $K_{\map,{j_\star}} \in \R^{n \times n_\star}$ and $K_{\map,\map} \in \R^{n \times n}$ are computed with eq.~\eqref{eq:nskernel} over the latent vectors $(\tbl_\map,\tbs_\map)$ over inputs $\x$, or using $(\tbl_{j_\star},\tbs_{j_\star})$ over inputs $\xs$. The simplest approximation is to choose only a single sample $s=1$ from the conditionals by choosing the conditional means $\tbl_{j_\star} = \mathbb{E}[\tbl_\star | \tbl_\map]$  and $\tbs_{j_\star} = \mathbb{E}[\tbs_\star | \tbs_\map]$ (See Supplementary Material). This is a sufficient approximation if the inputs $\x$ are sufficiently dense.

The predictive distribution given the HMC sample $\{ \tbl_i, \tbs_i, \tbo_i \}$ is derived analogously. We average over the $m$ HMC samples instead of a single MAP solution, and over the $s$ samples $\{ \tbl_{ij_\star} \}_{j=1}^s$  and $\{ \tbs_{ij_\star}  \}_{j=1}^s$  from the conditionals, resulting in
\begin{align} \label{eq:predhmc}
p(\fs | \y) \approx p(\fs | \{ \tbl_i, \tbs_i, \tbo_i \}, \y) &\approx \frac{1}{ms} \sum_{i=1}^m \sum_{j=1}^s p(\fs | \tbl_i, \tbl_{ij_\star}, \tbs_i, \tbl_{ij_\star}, \tbo_i, \y) \nonumber \\
 & \sim \frac{1}{ms} \sum_{i=1}^m \sum_{j=1}^s  \N(\bmu_{i,{j_\star}}, \Sigma_{i,{j_\star}} )
\end{align}
where $\bmu_{i,{j_\star}} = K_{i,{j_\star}}^T(K_i + \Omega_i)^{-1} \y$ and $\Sigma_{i,{j_\star}} = K_{{j_\star},{j_\star}} - K_{i,{j_\star}}^T(K_i + \Omega_i)^{-1} K_{i,{j_\star}}$, and where the kernel matrices are computed using $\tbl_i,\tbs_i$ and $\tbl_{ij_\star},\tbs_{ij_\star}$.

We note that a slower but perhaps more elegant alternative is to model latent functions jointly over  concatenated inputs $\x_t \equiv (\x,\xs)$, resulting in $\bl_t \equiv (\bl, \bl_\star)$, and analogously for the other functions. In this case the function posterior contains the predictive posterior, but the latent vector sizes increase to $n + n_\star$.

\section{Experiments}
\label{sec:exp}

We assess the performance of the proposed method on several synthetic and real datasets. We experiment with 8 synthetic datasets and a gene expression time series dataset \citep{heinonen15}. Of the synthetic datasets, three datasets are from the literature: the motorcycle dataset \textbf{M} \citep{silverman85}, the `jump' dataset \textbf{J} \citep{paciorek04} and a nonstationary dataset \textbf{T} from GPstuff (\texttt{demo\_epinf} in \citet{vanhatalo13}). We also generated five synthetic datasets with different types of nonstationarities (See Table \ref{tab:datasets}). We expect datasets exhibiting specific types of input-dependent characteristics to require a model with the corresponding input-dependencies.

\begin{table}[ht]
\centering
\caption{Datasets with varying forms of non-stationarities and sizes.}
\begin{tabular}{ccccc}
Dataset & Non-stationary functions & $n$ & $n_\text{train}$ & Comment \\
\hline
\textbf{M}$_\o$   & $\o(t)$ & 133 & 67 & Motorcycle dataset \citep{silverman85} \\
\textbf{D}$_\s$ & $\s(t)$  & 100 & 50 & \\
\textbf{D}$_\l$ & $\l(t)$  & 150 & 75 & \\
\textbf{D}$_{\s,\o}$ & $\s(t), \o(t)$  & 100 & 50 & \\
\textbf{D}$_{\l,\o}$ & $\l(t),\o(t)$  & 150 & 75 & \\
\textbf{D}$_{\l,\s,\o}$ & $\l(t),\s(t),\o(t)$  & 90 & 45 & \\ 
\textbf{J}    & N/A     & 101 & 50 & Jump dataset (3rd) \citep{paciorek04} \\
\textbf{T}$_{\s,\o}$    & $\o(t),\s(t)$     & 500 & 250 & from GPstuff \texttt{demo\_epinf} \citep{vanhatalo13}
\end{tabular}
\label{tab:datasets}
\end{table}

All dataset outputs are normalised to range $[-1, 1]$ and inputs to range $[0,1]$. For each dataset, we use half of the data as training data and the rest as test data. We will assess the performance against the test data with mean squared error $\text{MSE} = \frac{1}{n_{\text{test}}} \sum_i (y_i^{test} - [\bmu_\star]_i)^2$ and the negative log probability density $\text{NLPD} = \sum_i \log p(y_i^{test} | [\bmu_\star]_i , [\Sigma_\star]_{ii})$ values. For consistency, we model stationary parameters as vectors $c \mathbf{1}$ of length $n_{train}$ for $c = \{\ \o, \s, \l\}$. 

We run MAP optimisation from 10 different initial conditions and choose the one with the highest MLL value. We run 10 chains of 1000 samples of HMC-NUTS sampling using model whitening (Algorithm 3 of \citet{hoffman14}, $\epsilon = 0.01$, maximum tree depth $10$). The hyperparameters are set to $\alpha=1$, and $\mu_\l = 0.2$, $\mu_\s = 0.5$ and $\mu_\o = 0.1$. We set the $\beta$ parameters to sensible values of $\beta_\l = \beta_\s = 0.1$ and $\beta_\o = 0.2$.

\subsection{Regression performance}

Table \ref{tab:map} shows the MSE and NLPD performance on the test folds of the synthetic datasets using various MAP GP models. For each dataset, the model with the smallest NLPD, or second smallest NLPD value, is the one where the model's nonstationarities match those of the dataset. For instance, the dataset \textbf{T} contains heteroscedatic noise and input-dependent $\s$. For this, the best performance is obtained with a matching $\o,\s$-GP and with a fully nonstationary $\o,\s,\l$-GP as well. 

With MSE the stationary GP performs slightly better, albeit still worse than the optimal non-stationary method. This applies for every dataset. 

Adding `unnecessary' nonstationarities retains or only slightly worsens the performance, with the major exception being the `jump' dataset \textbf{J}. Here, the lengthscale is clearly input-dependent (NLPD $-0.72$, optimal), while in contrast the nonstationary signal variance $\s$ is unable to model the data (NLPD $1.04$). Adding Heteroscedatic noise to any of the models of this dataset weakens the model.

\begin{table}[t]
\centering
\caption{Test MSE and NLPD results on the synthetic datasets over various MAP models. Optimal values are in boldface. The optimal or second to optimal NLPD values follow a diagonal line. Smaller values are better for both quantities.}
\resizebox{\columnwidth}{!} {
\begin{tabular}{r|cc|cc|cc|cc|cc|cc|cc|cc|}
 & \multicolumn{2}{c|}{\textbf{M}$_\o$} & \multicolumn{2}{c|}{\textbf{D}$_{\s}$} & \multicolumn{2}{c|}{\textbf{D}$_{\l}$} & \multicolumn{2}{c|}{\textbf{D}$_{\o,\s}$} & \multicolumn{2}{c|}{\textbf{D}$_{\o, \l}$} & \multicolumn{2}{c|}{\textbf{D}$_{\o, \s, \l}$} & \multicolumn{2}{c|}{\textbf{J}} & \multicolumn{2}{c|}{\textbf{T}$_{\o,\s}$} \\
\cline{2-17}
Method & MSE & NLPD & MSE & NLPD & MSE & NLPD & MSE & NLPD & MSE & NLPD & MSE & NLPD & MSE & NLPD & MSE & NLPD \\
\hline
GP            & $3.91$ & $0.11$ & $0.21$ & $-1.31$ & $0.71$ & $-0.85$ & $0.44$ & $-0.86$ & $0.54$ & $-1.04$ & $0.16$ & $-1.25$ & $1.48$ & $-0.32$ & $17.83$ & $0.10$ \\ 
$(\o)$-GP     & $3.91$ & $\mathbf{-0.22}$ & $0.21$ & $-1.27$ & $2.46$ & $-0.47$ & $0.45$ & $-0.98$ & $2.69$ & $-1.28$ & $0.33$ & $-1.73$ & $3.51$ & $-0.58$ & $17.35$ & $0.01$ \\ 
$(\s)$-GP     & $3.93$ & $0.17$ & $\mathbf{0.18}$ & $-1.37$ & $0.64$ & $-0.93$ & $0.43$ & $-0.94$ & $0.52$ & $-1.07$ & $0.17$ & $-0.42$ & $1.38$ & $1.04$ & $16.84$ & $0.07$ \\ 
$(\l)$-GP     & $3.94$ & $0.16$ & $\mathbf{0.18}$ & $\mathbf{-1.38}$ & $\mathbf{0.53}$ & $\mathbf{-1.05}$ & $0.44$ & $-0.88$ & $0.41$ & $-1.16$ & $0.17$ & $-0.34$ & $\mathbf{1.35}$ & $\mathbf{-0.72}$ & $17.63$ & $0.09$ \\ 
$(\o,\s)$-GP   & $\mathbf{3.87}$ & $\mathbf{-0.23}$ & $\mathbf{0.18}$ & $-1.30$ & $0.65$ & $-0.86$ & $0.43$ & $\mathbf{-1.07}$ & $0.56$ & $-1.51$ & $\mathbf{0.15}$ & $-1.61$ & $1.60$ & $-0.66$ & $16.33$ & $\mathbf{-0.02}$ \\ 
$(\o,\l)$-GP   & $4.02$ & $-0.19$ & $0.19$ & $-1.31$ & $\mathbf{0.53}$ & $-0.93$ & $\mathbf{0.42}$ & $-0.99$ & $\mathbf{0.40}$ & $\mathbf{-1.83}$ & $0.17$ & $-0.90$ & $1.38$ & $-0.47$ & $\mathbf{9.30}$ & $0.01$ \\ 
$(\o,\s,\l)$-GP & $3.90$ & $-0.21$ & $0.19$ & $-1.32$ & $\mathbf{0.53}$ & $-0.90$ & $0.45$ & $-0.98$ & $0.43$ & $-1.70$ & $0.16$ & $\mathbf{-1.79}$ & $1.47$ & $-0.32$ & $9.77$ & $-0.00$ \\ 
\end{tabular}
}
\label{tab:map}
\end{table}

\subsection{HMC performance}

\begin{figure}[t]
\centering
\subfloat[All datasets and models]{ \includegraphics[width=0.47\columnwidth]{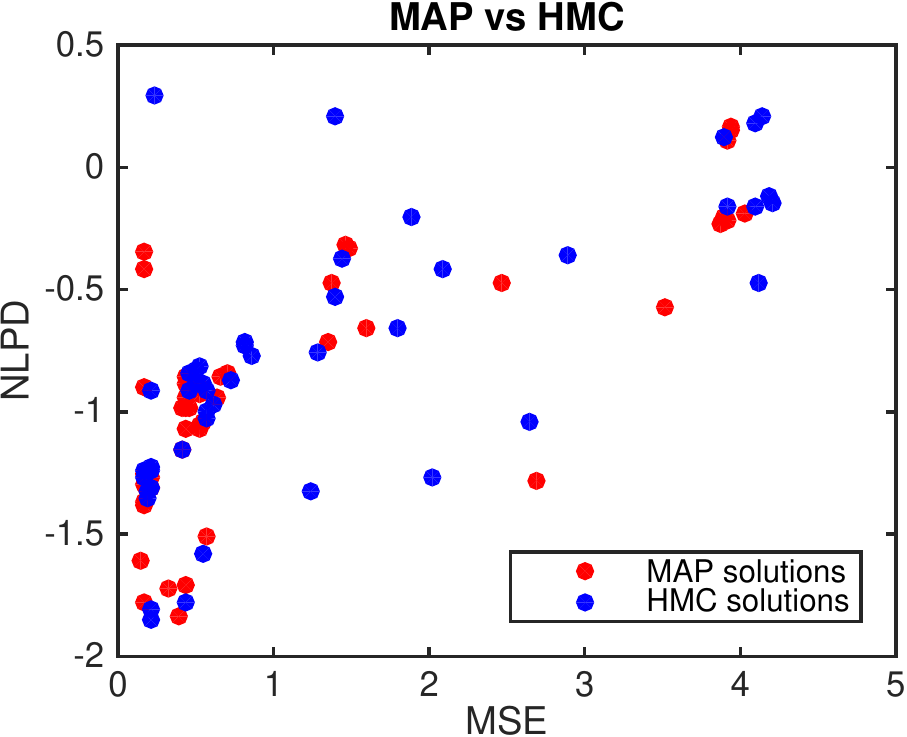} }
\subfloat[\textbf{D}$_\l$ dataset with $\l$-GP model]{ \includegraphics[width=0.47\columnwidth]{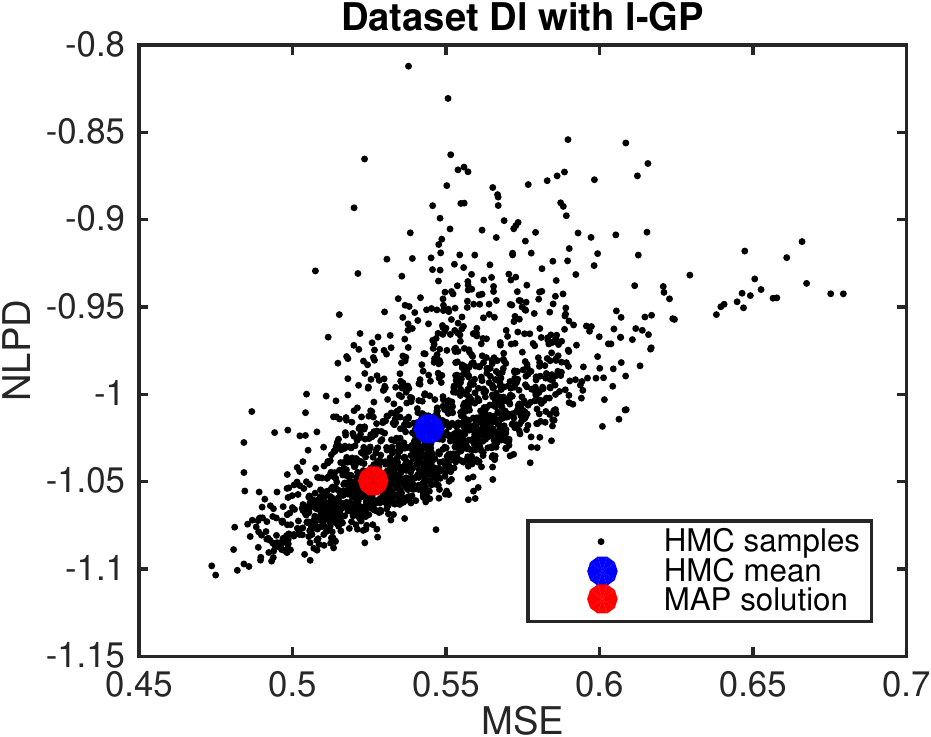} }
\caption{The MSE and NLPD performance of the HMC posterior samples. (a) Comparison of test errors between MAP and HMC mean solutions over all datasets and methods ($8 \times 7 = 56$, x-axis limited to $5$ for clarity). (b) Test errors of the HMC samples compared to the HMC mean and MAP solution on a single $D_\l$ dataset with $\l$-GP model.}
\label{fig:hmc}
\end{figure}

We explored the difference between the MAP solution and the HMC sampling. In practice we found the MAP to be slightly better on average regarding the MSE and NLPD values (See Figure \ref{fig:hmc}a). However, the sampling solution is robust against multimodality in the latent posterior. Figure \ref{fig:hmc}b shows the test errors of the individual HMC samples in comparison to the MAP solution with the \textbf{D}$_\l$ dataset using the $\l$-GP model. The HMC solution includes numerous samples that are better, while on average being slightly worse than MAP. 

\begin{figure}[ht]
\centering
\includegraphics[width=0.99\columnwidth]{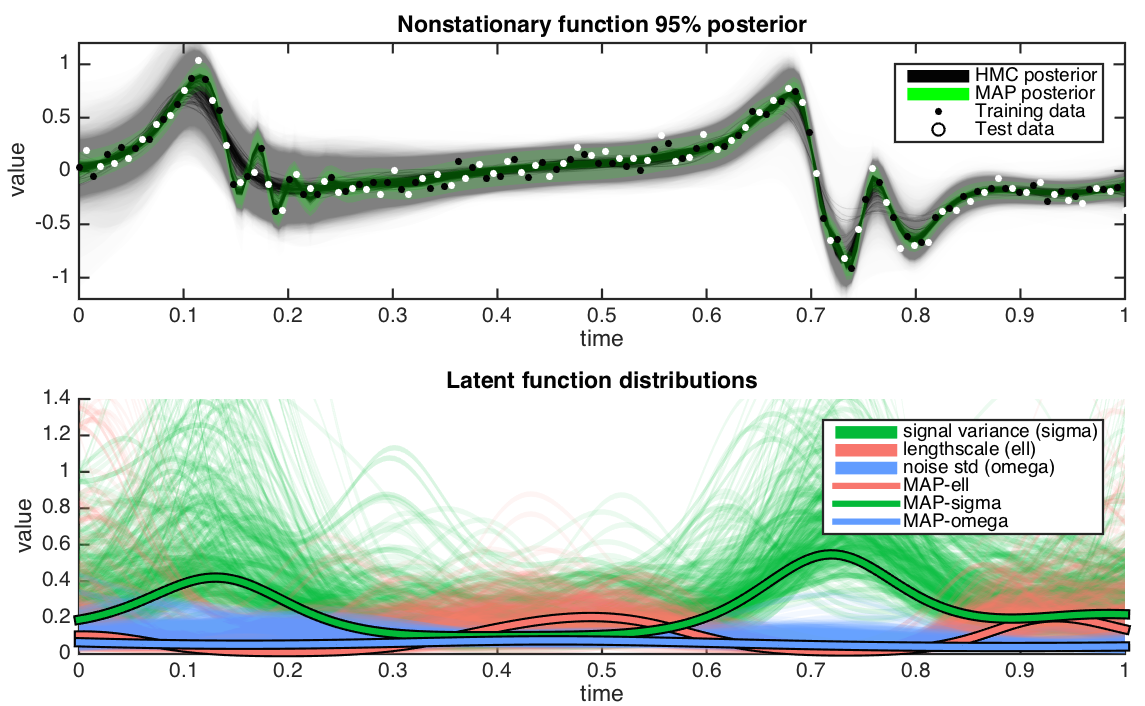} 
\caption{The set of function posteriors corresponding to the latent function samples are plotted in gray, and the MAP solution in green, along with the training and test data (top). The latent function sample is drawn with green ($\s$), red ($\l$) and blue ($\o$) colors, with the MAP solution highlighted with bold lines (bottom).}
\label{fig:example}
\end{figure}

The dataset \textbf{D}$_\l$ contains several latent modes (See Figure \ref{fig:example}, bottom), which the HMC sampler captures. These modes include latent functions that imply a `shortcut' or a `zigzag' signal around timepoints $0.18$ or $0.75$, or both. The HMC samples are centered mostly around the shortcut profile at the earlier timepoint, while only a few samples with the shortcut profile exist at the later timepoint. The MAP solution has chosen both zigzags. The latent posterior shows largest variance in the signal variance $\s(t)$ component, while the lengthscale $\l(t)$ and noise variance $\o$ have tighter distributions.

\subsection{Biological dataset}

We showcase the method with a biological dataset of $205$ gene expression time series measurements of human endothelial cells after irradiation at time $t=0$. Due to the irradiation the dataset exhibits nonstationary dynamics as the cells try to repair themselves and revert back to steady states. The gene expressions are measured over 8 days $(0.5, 1, 2, 3, 4, 7, 14, 21)$ in three replicates \citep{heinonen15}. The goal is to construct a realistic model of the underlying gene expression process and the underlying dynamics with no knowledge of the `true' expression levels, given only the small number of sparse measurements.

\begin{figure}[ht]
\centering
\subfloat{ \includegraphics[width=0.33\columnwidth]{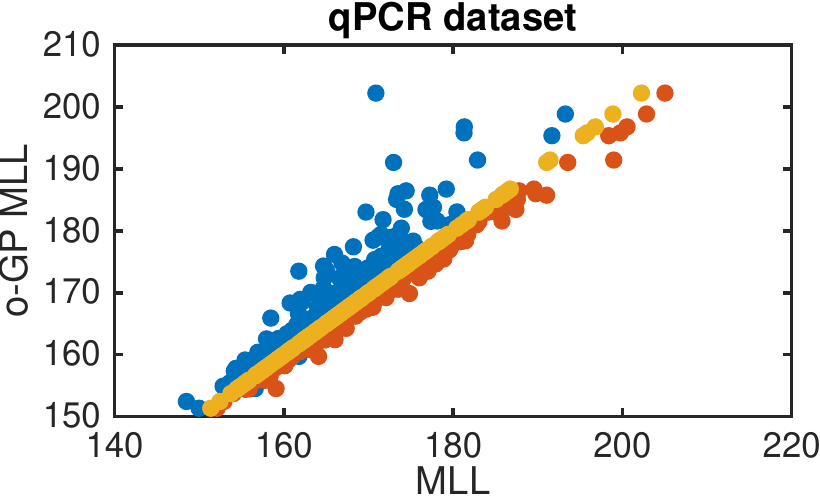} }
\subfloat{ \includegraphics[width=0.33\columnwidth]{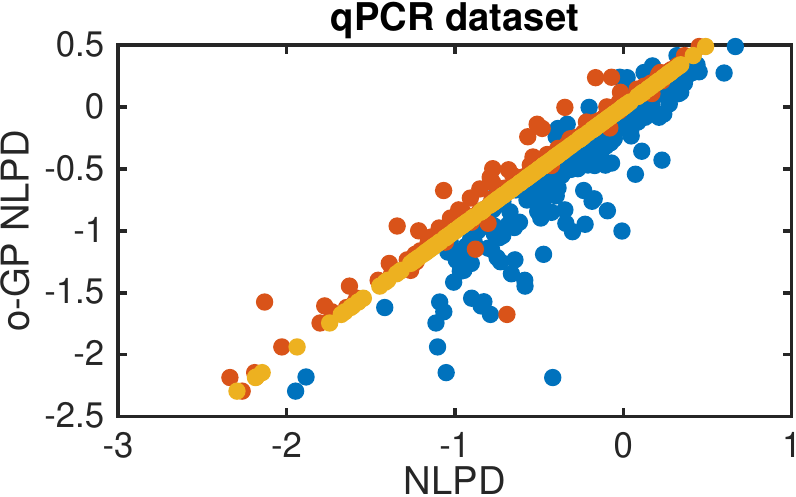} }
\subfloat{ \includegraphics[width=0.33\columnwidth]{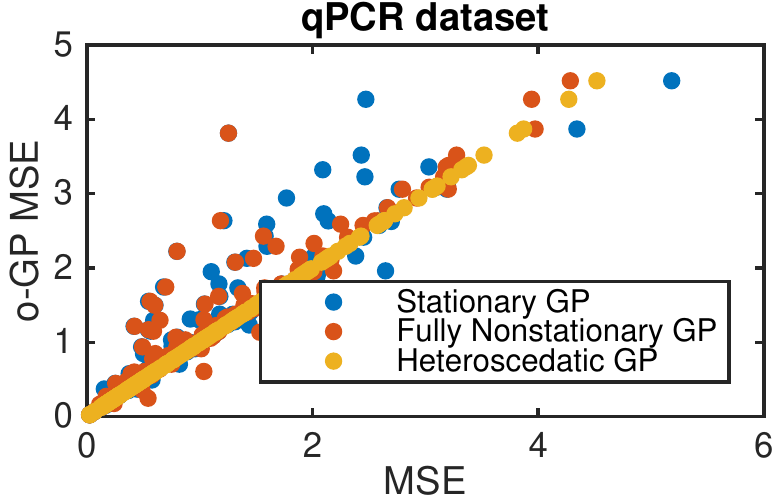} } \\
\caption{Comparison of the MLL, MSE and NLPD values (x-axis) of the stationary GP, $\o$-GP and $(\o,\s,\l)$-GP over the 205 gene expression time series (x-axis) against the heteroscedastic GP on the y-axis. Each row contains a triplet of values corresponding to the three GP models of the same time series.}
\label{fig:qpcr}
\end{figure}

\begin{figure}[ht]
\centering
\subfloat{ \includegraphics[width=0.33\columnwidth]{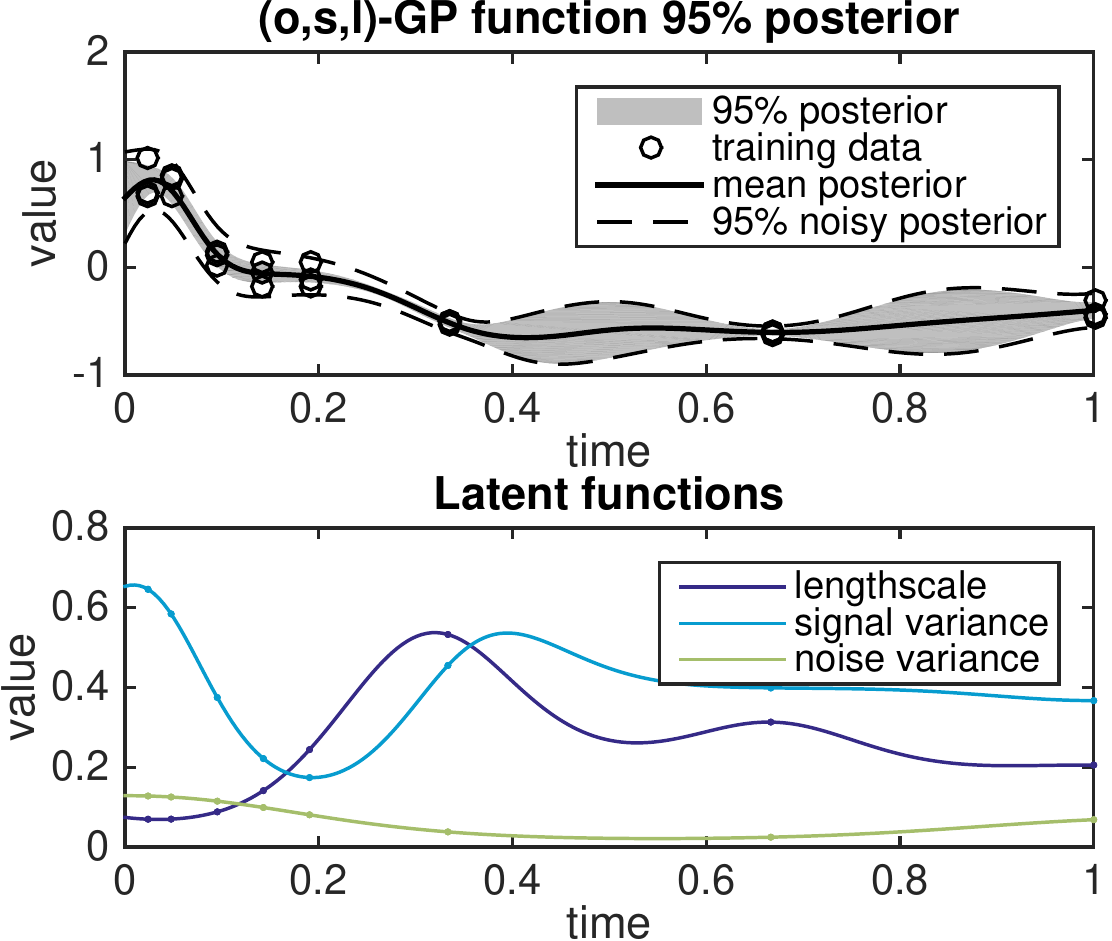} }
\subfloat{ \includegraphics[width=0.33\columnwidth]{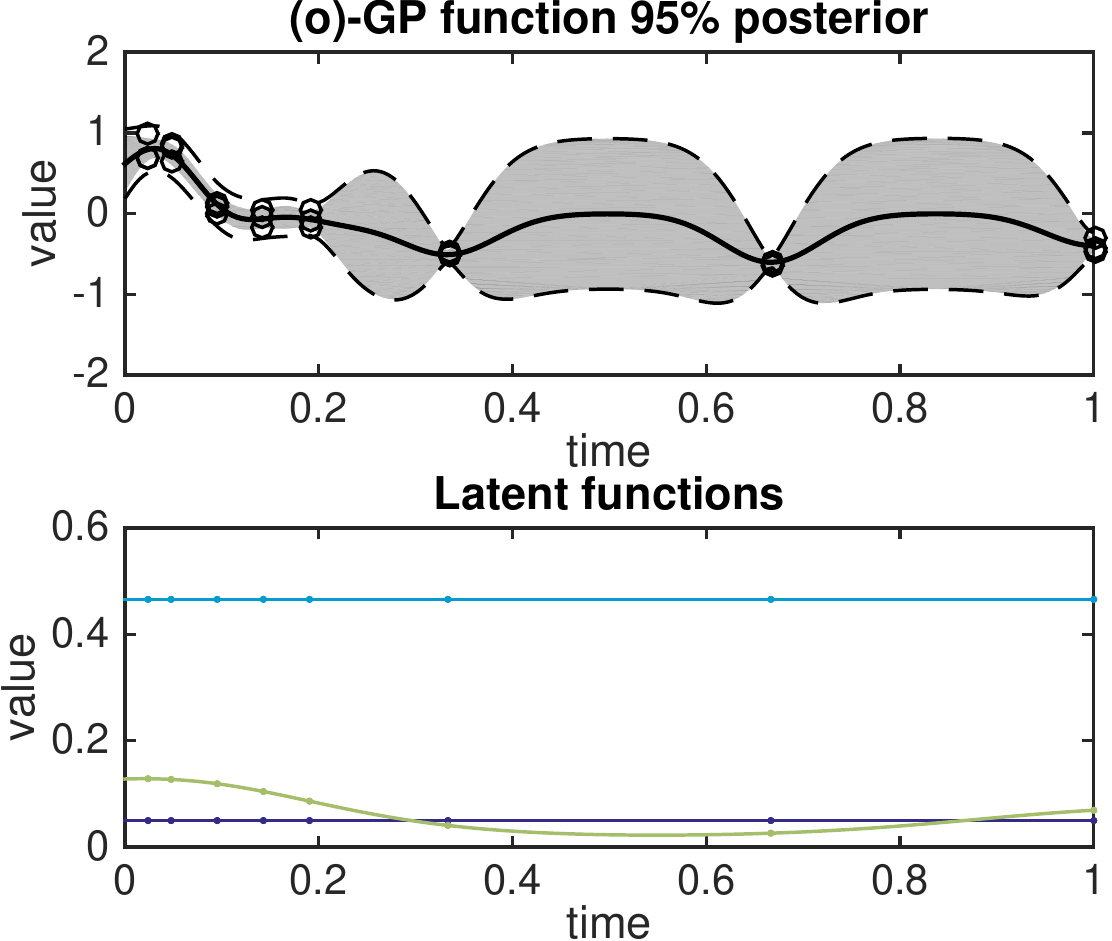} }
\subfloat{ \includegraphics[width=0.33\columnwidth]{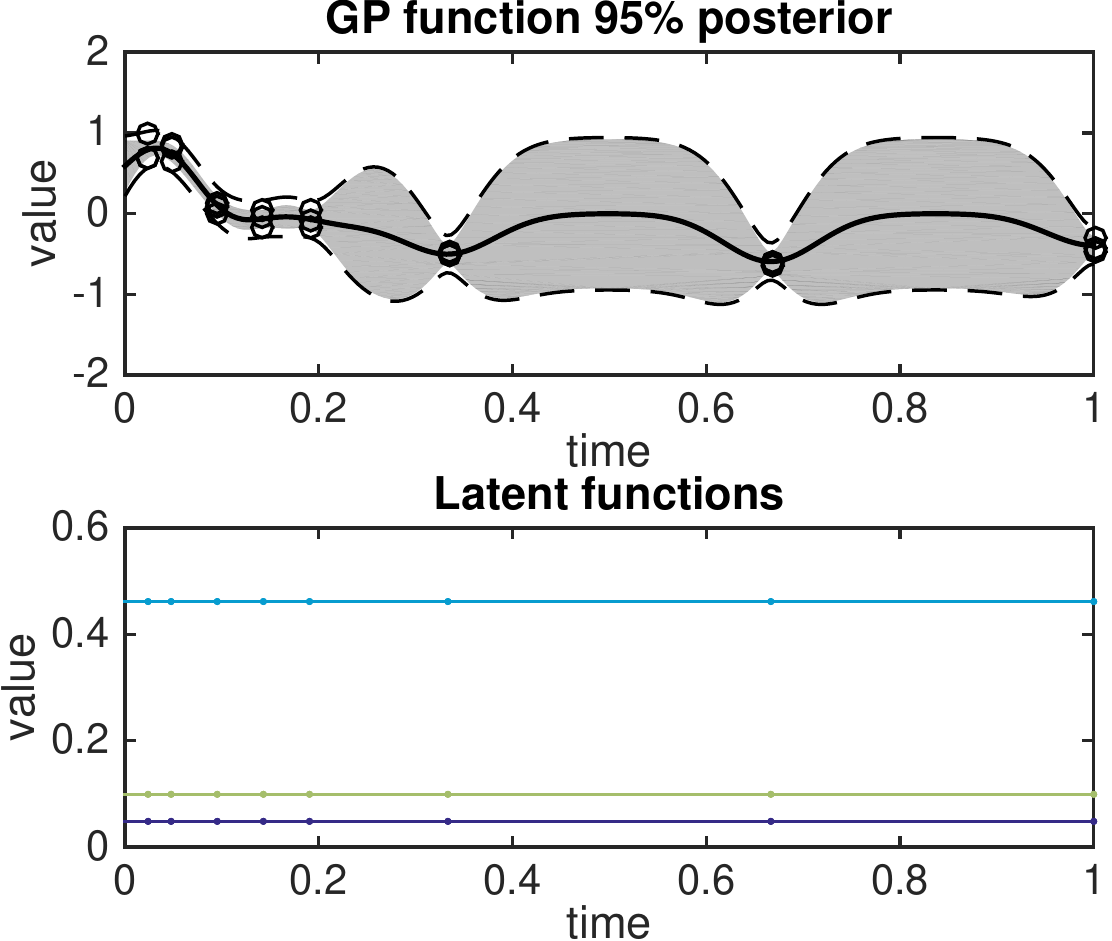} } \\
\caption{Comparison of three GP models on an example expression time series.}
\label{fig:qpcr140}
\end{figure}

We modeled the dataset using stationary GP, heteroscedatic $\o$-GP and three nonstationary GPs: $(\o,\s)$-GP, $(\o,\l)$-GP and with $(\o,\s,\l)$-GP. We found the performance of the three nonstationary GPs to be similar. Figure \ref{fig:qpcr} indicates the MLL, MSE and NLPD values of the 205 timeseries under stationary, heteroscedatic or fully nonstationary models. Addition of heteroscedasticity greatly increases the model fits, while also improving the data likelihoods against the function posterior. Finally the fully nonstationary GP still improves model fits, while consistently improving the NLPD values, with similar MSE performance compared to the HGP. Figure \ref{fig:qpcr140} compares the three models learned from an an example gene expression time series.



\subsection{Latent function reconstruction}

\begin{figure}
     \centering
     \subfloat[Latent errors]{\includegraphics[width=0.67\columnwidth]{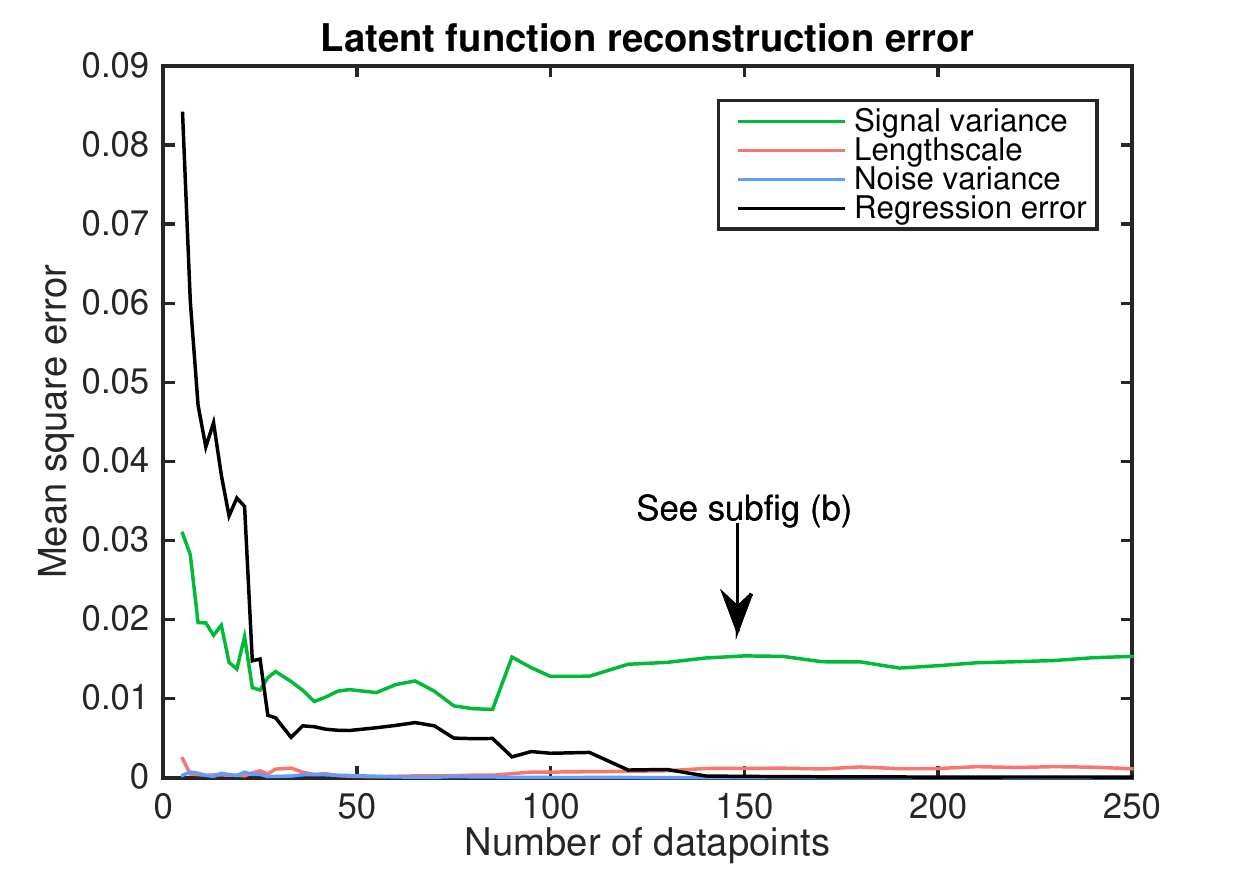}} \\ \vskip 3mm
     \subfloat[150 data points]{\includegraphics[width=0.65\columnwidth]{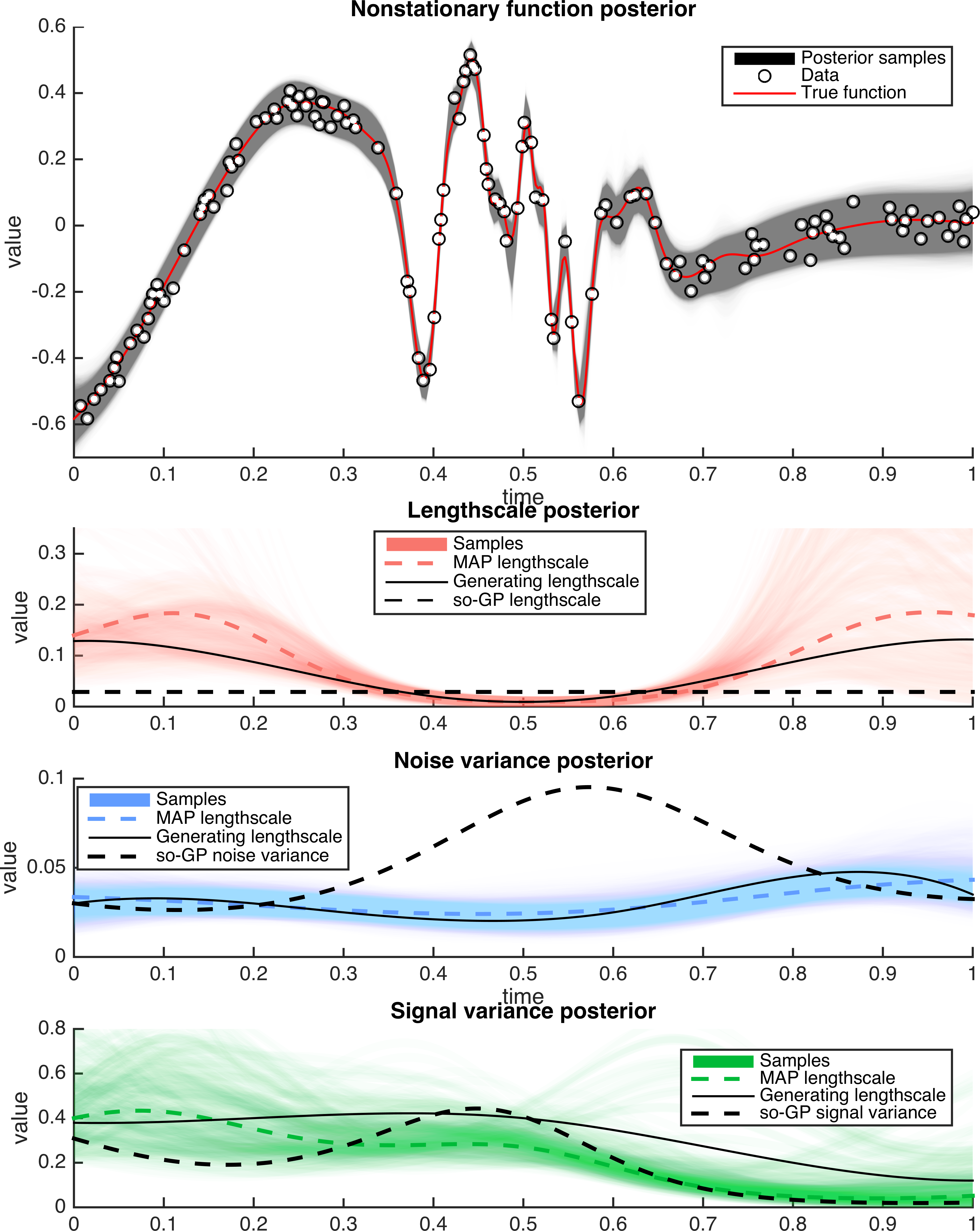}}
     \caption{Latent function reconstruction errors (top) over a nonstationary simulated dataset with known nonstationary dynamics. Fully nonstationary GP model (bottom) posteriors, MAP solution (dashed line) and generating latent functions (black solid lines) with $150$ data points. The latent lengthscales and noises are estimated correctly, while signal variance is approximately matched. Dashed black lines shows the comparison to the state-of-the-art $\sigma,\omega$-GP model of \citet{tolvanen14}.}
     \label{fig:reconstruction}
\end{figure}

Finally, we highlight the methods potential in reconstructing the latent parameter processes. We simulate a noisy sample where true generating latent parameter functions $\l(\cdot),\s(\cdot),\o(\cdot)$ are known. We computed both the MAP solution and sampled the posterior of the latent space and of the unknown function $f(\cdot)$ using HMC inference and showcase the results in Figure \ref{fig:reconstruction}. The latent function reconstruction error curve against increasing numbers of noisy points shows that the regression error, and lengthscale and noise variances converge to true values (Figure \ref{fig:reconstruction} top), while the signal variance shows small bias but is still well estimated. 

Figure \ref{fig:reconstruction} bottom highlights the MAP and HMC solutions given $150$ datapoints, and compares them to the state-of-the-art $\sigma,\omega$-GP model of \citet{tolvanen14}. We are able to accurately estimate the latent functions.

\section{Discussion}
\label{sec:conclusions}

In this paper, we have proposed a fully non-stationary Gaussian process regression framework, where all three key components can be input-dependent. Our approach uses analytical gradient-based techniques to perform inference with HMC sampling and MAP estimation. We are able to effectively sample from the exact posterior of the latent functions. We have shown that the method is able to infer the underlying latent functions and improve regression performance when the datasets truly are nonstationary, and achieve equivalent performance to a stationary model when they are not.

The interplay between the signal variance and the lengthscale is an interesting topic \citep{diggle98,zhang04}.  When modeling the `jump' dataset the signal variance was unable to model dynamics, while nonstationary lengthscale produced a good model. This is natural since the signal variance serves as a linear amplitude, while the lengthscale has a possibly non-linear effect on the function model.

The gradient-based HMC is a powerful inference tool for Gaussian processes, and could be further enhanced by utilizing natural gradients or position-dependent mass matrices with Riemannian Manifold HMC \citep{girolami11}. We note that the method could be extended by also inferring the hyperparameters $\bt$ using HMC. However, proper care has to be taken to set their priors.

\clearpage

\section*{Supplemental information}

\subsection*{Comparison between vanilla and $(\o,\s,\l)$-GP} 

A comparison between stationary (vanilla) and $(\o,\s,\l)$-GP is shown in Supplemental Figure \ref{fig:example2}.

\begin{figure}[h]
     \centering
     \subfloat[Vanilla GP (NLPD $-1.95$)]{\includegraphics[width=0.49\columnwidth]{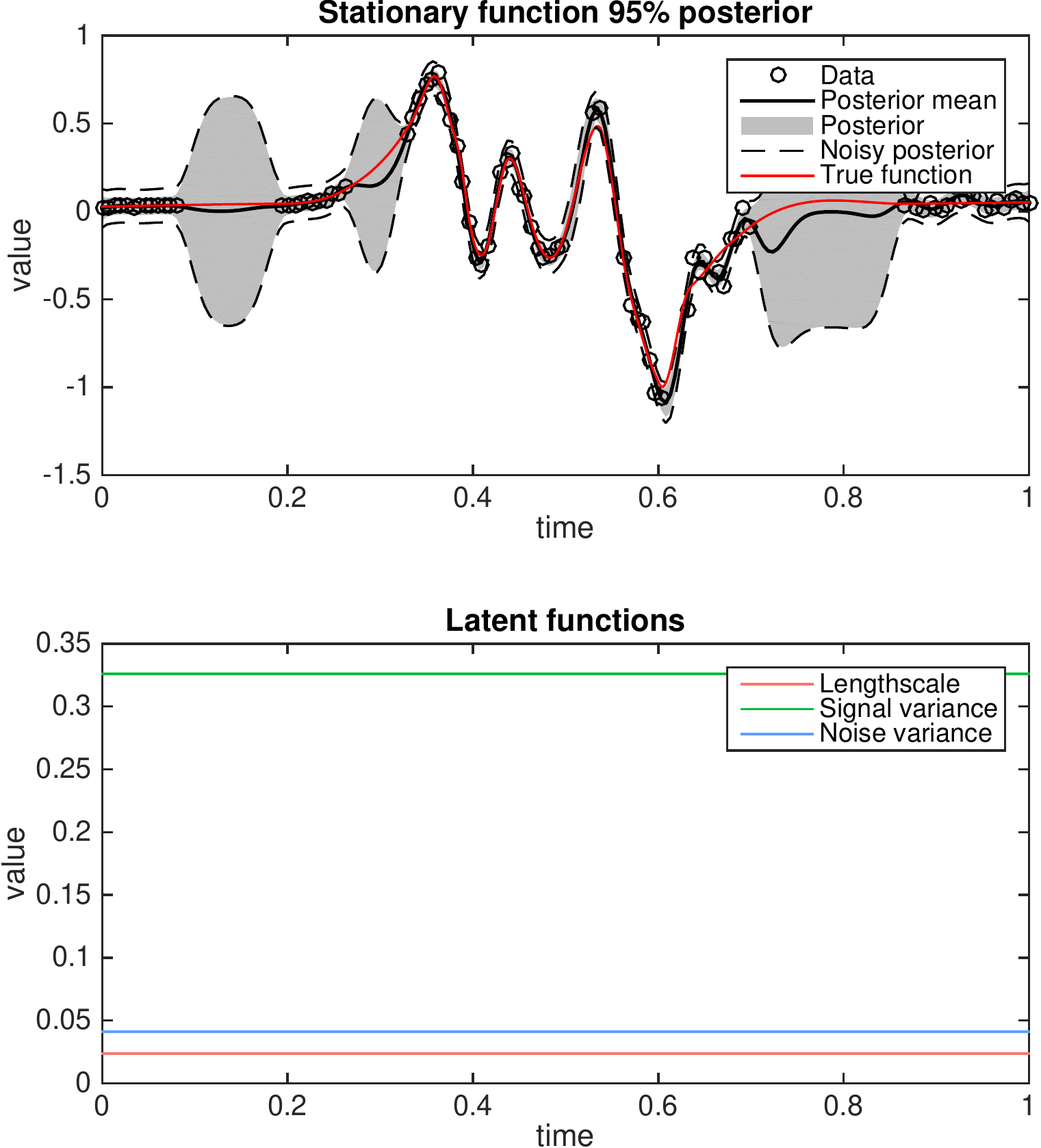}}
     \subfloat[Fully nonstationary GP (NLPD $-2.47$)]{\includegraphics[width=0.49\columnwidth]{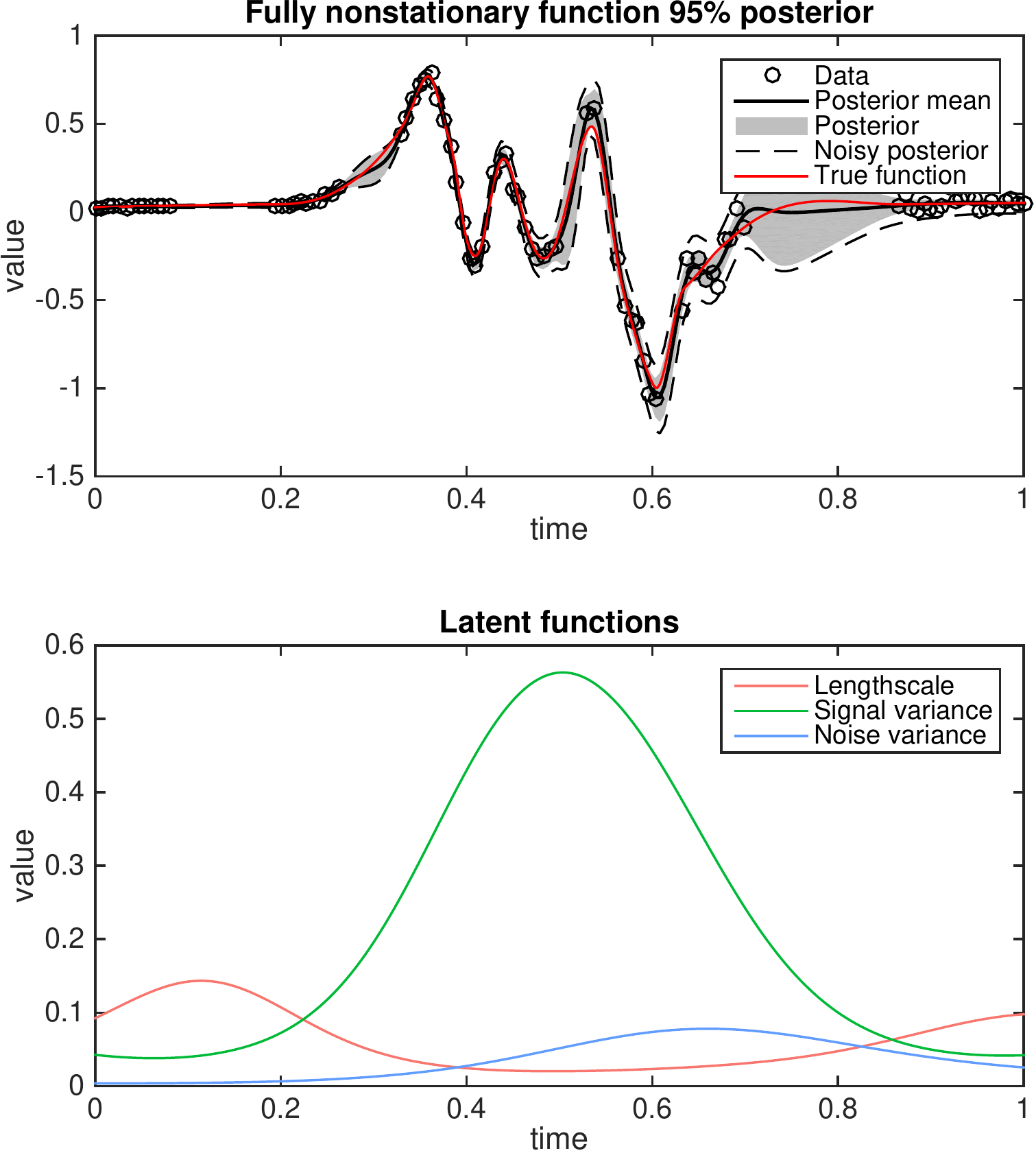}}
     \caption{Nonstationary, heteroscedastic GP (right) fits nonstationary dataset better than vanilla GP (left). Vanilla GP overestimates confidence intervals at stationary regions, as a result of having to choose the lengthscale to fit the nonstationary regions, and for the same reason produces spurious oscillations in regions where there are no observations. The dataset is the $\mathbf{D}_{\o,\s,\l}$ from Table 2.}
      \label{fig:example2}
\end{figure}

\subsection*{Kernel SDP proof}

The kernel 
$$K(x,x') = \sqrt{\frac{2 \l(x) \l(x')}{ \l(x)^2 + \l(x')^2 }} \exp \left( - \frac{(x - x')^2}{\l(x)^2 + \l(x')^2}  \right)$$
is a one-dimensional Gaussian SDP kernel $C^{NS}(\x_i, \x_j)$ of \citet{paciorek04}. The kernel $K_f$ can be stated as
$$K_f(x,x') = \s(x) K(x,x') \s(x'),$$
which is positive definite for any function $\s(\cdot)$ \citep{shawetaylor04}.

\subsection*{Conditional distributions}

The conditional distributions of the latent functions at target timepoints $\xs$ given the latent functions at observed timepoints $\x$ are
\begin{align*}
p(\tbl_\star | \tbl) &= \N(\tbl_\star |  K_\l(\xs, \x)^T  K_\l(\x, \x)^{-1} (\tbl - \mu_\l) + \mu_\l, K_\l(\xs, \xs) - K_\l(\xs, \x)^T K_\l(\x, \x)^{-1} K_\l(\xs, \x) ) \\ 
p(\tbs_\star | \tbs) &= \N(\tbs_\star |  K_\s(\xs, \x)^T  K_\s(\x, \x)^{-1} (\tbs - \mu_\s) + \mu_\s, K_\s(\xs, \xs) - K_\s(\xs, \x)^T K_\s(\x, \x)^{-1} K_\s(\xs, \x) ) \\ 
p(\tbo_\star | \tbo) &= \N(\tbo_\star |  K_\o(\xs, \x)^T  K_\o(\x, \x)^{-1} (\tbo - \mu_\o) + \mu_\o, K_\o(\xs, \xs) - K_\o(\xs, \x)^T K_\o(\x, \x)^{-1} K_\o(\xs, \x) ),
\end{align*}
where $K_\l$, $K_\s$ and $K_\o$ are standard gaussian kernels computed using the hyperparameters $\bt$ \citep{rasmussen06}.

\subsection*{Partial derivatives}

The partial derivatives of the unconstrained latent functions against the marginal log likelihood 
\begin{align*}
\log \L = \log p(\y | \tbl, \tbs, \tbo, \theta) &= \log p(\y | \tbl, \tbs, \tbo) p(\tbl | \theta) p(\tbs | \theta) p(\tbo | \theta) \\
  &= \log \N(\y | \0, K_f + \Omega) + \log \N(\tbl | \mu_\tl, K_\tl) + \log \N(\tbs | \mu_\ts, K_\ts) + \log \N(\tbo | \mu_\to, K_\to) 
\end{align*}
are analytically derived. 

The partial derivative for $\tl(t)$ is
\begin{align*}
\frac{\partial \log \L}{\partial \tbl_i }  &= \frac{1}{2} \tr \left( (\balpha \balpha^T - K_y^{-1}) \frac{\partial K_y}{\partial \tbl_i} \right) - [K_{\tl}^{-1} (\tbl - \mu_{\tl})]_i \\
\frac{\partial [K_y]_{ij}}{\partial \tbl_i }  &= \frac{S_{ij} E_{ij}}{R_{ij} L_{ij}^3} \l_i \l_j (4 d \l_i^2 - \l_i^2 + \l_j^4),
\end{align*}
$S_{ij} = \s_i \s_j$, $R_{ij} = \sqrt{\frac{2\l_i \l_j}{\l_i^2 + \l_j^2}}$, $E_{ij} = \exp\left( - \frac{(t_i - t_j)^2}{\l_i^2 + l_j^2} \right)$, and $L_{ij} = \l_i^2 + \l_j^2$. The derivative matrix $\frac{\partial K_y}{\partial \tbl_i}$ becomes a 'plus' matrix where only $i$'th column and row are nonzero.

The derivatives for $\ts(t)$ is
\begin{align*}
\frac{\partial \log \L}{\partial \tbs } &= \diag \left( (\balpha \balpha^T - K_y^{-1}) K_f \right) - K_{\ts}^{-1} (\tbs - \mu_{\ts}) 
\end{align*}
and for $\to(t)$ is
\begin{align*}
\frac{\partial \log  \L}{\partial \tbo }  &= \diag \left( (\balpha \balpha^T - K_y^{-1}) \Omega \right) - K_{\to}^{-1} (\tbo - \mu_{\to})
\end{align*}	
where $\balpha = K_y^{-1} \y$.

The partial derivatives of the latent parameters $(\tl,\ts,\to) \in \R^{3}$ in a stationary formulation are for $\tl$ \citep{rasmussen06}
\begin{align*}
\frac{\partial \L}{\partial \tl }  &= \frac{1}{2} \tr \left( (\balpha \balpha^T - K_y^{-1}) \frac{\partial K_y}{\partial \tl} \right) - \mathbf{1}^T K_{\tl}^{-1} (\tl \mathbf{1} - \mu_{\tl}) \\
\frac{\partial K_y}{\partial \tl }  &= \l^{-2} D \odot K_f,
\end{align*}
for $\ts$ 
\begin{align*}
\frac{\partial \L}{\partial \ts }  &= \tr \left( (\balpha \balpha^T - K_y^{-1}) K_f \right) - \mathbf{1}^T K_{\ts}^{-1} (\ts \mathbf{1} - \mu_{\ts})
\end{align*}
and for $\to$
\begin{align*}
\frac{\partial \L}{\partial \to }  &= \tr \left( (\balpha \balpha^T - K_y^{-1}) \Omega \right) - \mathbf{1}^T K_{\to}^{-1} (\to \mathbf{1} - \mu_{\to}).
\end{align*}

\bibliographystyle{plainnat}
\bibliography{refs.bib}


\end{document}